%% file: main.tex
\documentclass[letterpaper]{article} 
\usepackage{aaai2026}  
\usepackage{times}  
\usepackage{helvet}  
\usepackage{courier}  
\usepackage[hyphens]{url}  
\usepackage{graphicx} 
\usepackage{natbib}  
\usepackage{caption} 
\usepackage{algorithm}
\usepackage{algorithmic}
\usepackage{subfigure}

\usepackage{url}            
\usepackage{booktabs}       
\usepackage{amsfonts}       
\usepackage{nicefrac}       
\usepackage{microtype}      
\usepackage{xcolor}         

\usepackage{subfigure}
\usepackage{caption}
\usepackage{graphicx}
\usepackage{float} 
\usepackage{subcaption}
\usepackage{amsmath}
\usepackage{multirow}  
\usepackage{enumitem}
\usepackage{colortbl}
\definecolor{customgreen}{RGB}{0,150,0}
\definecolor{rouse}{RGB}{244,177,131}
\usepackage[export]{adjustbox} 

\usepackage{newfloat}
\usepackage{listings}
\DeclareCaptionStyle{ruled}{labelfont=normalfont,labelsep=colon,strut=off} 
\floatstyle{ruled}
\newfloat{listing}{tb}{lst}{}
\floatname{listing}{Listing}
\title{Persistent Autoregressive Mapping with Traffic Rules for \\ Autonomous Driving}
\author{
    Shiyi Liang\textsuperscript{\rm 1,2}\thanks{Equal contribution.}\thanks{Work done during the internship at Amap, Alibaba Group.},
    Xinyuan Chang\textsuperscript{\rm 3}\footnotemark[1],
    Changjie Wu\textsuperscript{\rm 3}\footnotemark[1],
    Huiyuan Yan\textsuperscript{\rm 1,2},
    Yifan Bai\textsuperscript{\rm 5}, \\
    Xinran Liu\textsuperscript{\rm 3},
    Hang Zhang\textsuperscript{\rm 3},
    Yujian Yuan\textsuperscript{\rm 4},
    Shuang Zeng\textsuperscript{\rm 1,2},
    Mu Xu\textsuperscript{\rm 3},
    Xing Wei\textsuperscript{\rm 1,2}\thanks{Corresponding author: Xing Wei.}
}
\affiliations{
    \textsuperscript{\rm 1}State Key Laboratory of Human-Machine Hybrid Augmented Intelligence, Xi'an Jiaotong University \\
    \textsuperscript{\rm 2}School of Software Engineering, Xi'an Jiaotong University \\
    \textsuperscript{\rm 3}Amap, Alibaba Group\\
    \textsuperscript{\rm 4} The Hong Kong University of Science and Technology \\
    \textsuperscript{\rm 5} DAMO Academy, Alibaba Group\\
     sy\_liang2023@stu.xjtu.edu.cn, 
    \{changxinyuan.cxy, wuchangjie.wcj\}@alibaba-inc.com, weixing@mail.xjtu.edu.cn}

\usepackage{bibentry}
\begin{document} 
\maketitle
\thispagestyle{plain}
\begin{figure}[b]

\end{figure}
\input{sec/0_abstract}

\input{sec/1_intro}
\input{sec/2_related}
\input{sec/4_E2E-ARM}
\input{sec/5_experiments}

\input{sec/6_conclusion}

\section*{Acknowledgments}
This work was support by the National Natural Science Foundation of China No. 62572385, the Fundamental Research Funds for the Central Universities No. xxj032023020, and CAAI-CANN Open Fund, developed on OpenI Community.

\bibliography{aaai2026}

\end{document}

%% file: sec/0_abstract.tex
\begin{abstract}

    Safe autonomous driving requires both accurate HD map construction and persistent awareness of traffic rules, even when their associated signs are no longer visible. However, existing methods either focus solely on geometric elements or treat rules as temporary classifications, failing to capture their persistent effectiveness across extended driving sequences. In this paper, we present \textbf{PAMR} (\textbf{P}ersistent \textbf{A}utoregressive \textbf{M}apping with Traffic \textbf{R}ules), a novel framework that performs autoregressive co-construction of lane vectors and traffic rules from visual observations. Our approach introduces two key mechanisms: \textbf{Map-Rule Co-Construction} for processing driving scenes in temporal segments, and \textbf{Map-Rule Cache} for maintaining rule consistency across these segments. To properly evaluate continuous and consistent map generation, we develop MapDRv2, featuring improved lane geometry annotations. Extensive experiments demonstrate that PAMR achieves superior performance in joint vector-rule mapping tasks, while maintaining persistent rule effectiveness throughout extended driving sequences.

\end{abstract}

\begin{links}
\link{Code}{https://miv-xjtu.github.io/PAMR/}
\end{links}


%% file: sec/1_intro.tex

\section{Introduction}

Driving by the rules is fundamental to safe autonomous navigation. Inherently sequential in nature, driving requires continuous interpretation and application of traffic rules along the vehicle's trajectory. While existing High-Definition (HD) map construction ~\cite{liao2023maptr, maptrv2, chen2024maptracker, ben2022toponet} focuses on geometric elements like lane topology, it often overlooks traffic rules—semantic elements that govern driving behavior and persist beyond their signs' visibility. These rules and road geometry form an intrinsically interwoven, rule-governed space. Our core idea is that robust autonomous navigation demands \textbf{Persistent Driving by the Rules}, acknowledging these semantic guidelines' continuous influence across time and space, from initial observation through extended trajectories.

\begin{figure}[t]
    \centering
    \includegraphics[width=1.0\linewidth]{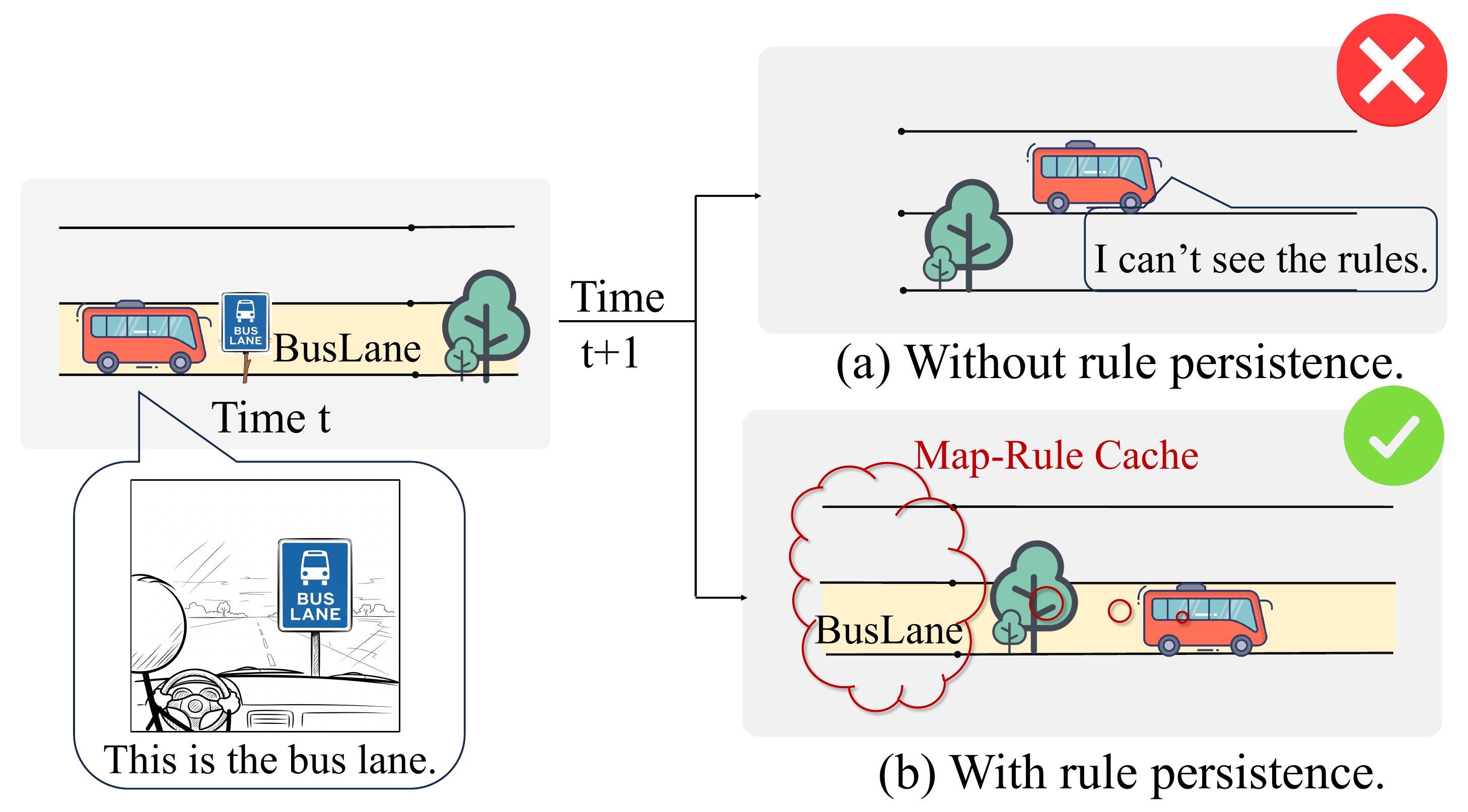} 
    \caption{Schematic of Persistent Rule Effectiveness. (a) \textbf{Without rule persistence}: At time t, a bus lane sign is observed. At t+1, without remembering the rule, the vehicle attempts an illegal lane change. (b) \textbf{With rule persistence:} The system retains the bus lane rule, preventing the incorrect maneuver and ensuring continued adherence to traffic rules.}
    \label{fig:traffic}
\end{figure}

\begin{figure}[t]
    \centering
    \includegraphics[width=1.0\linewidth]{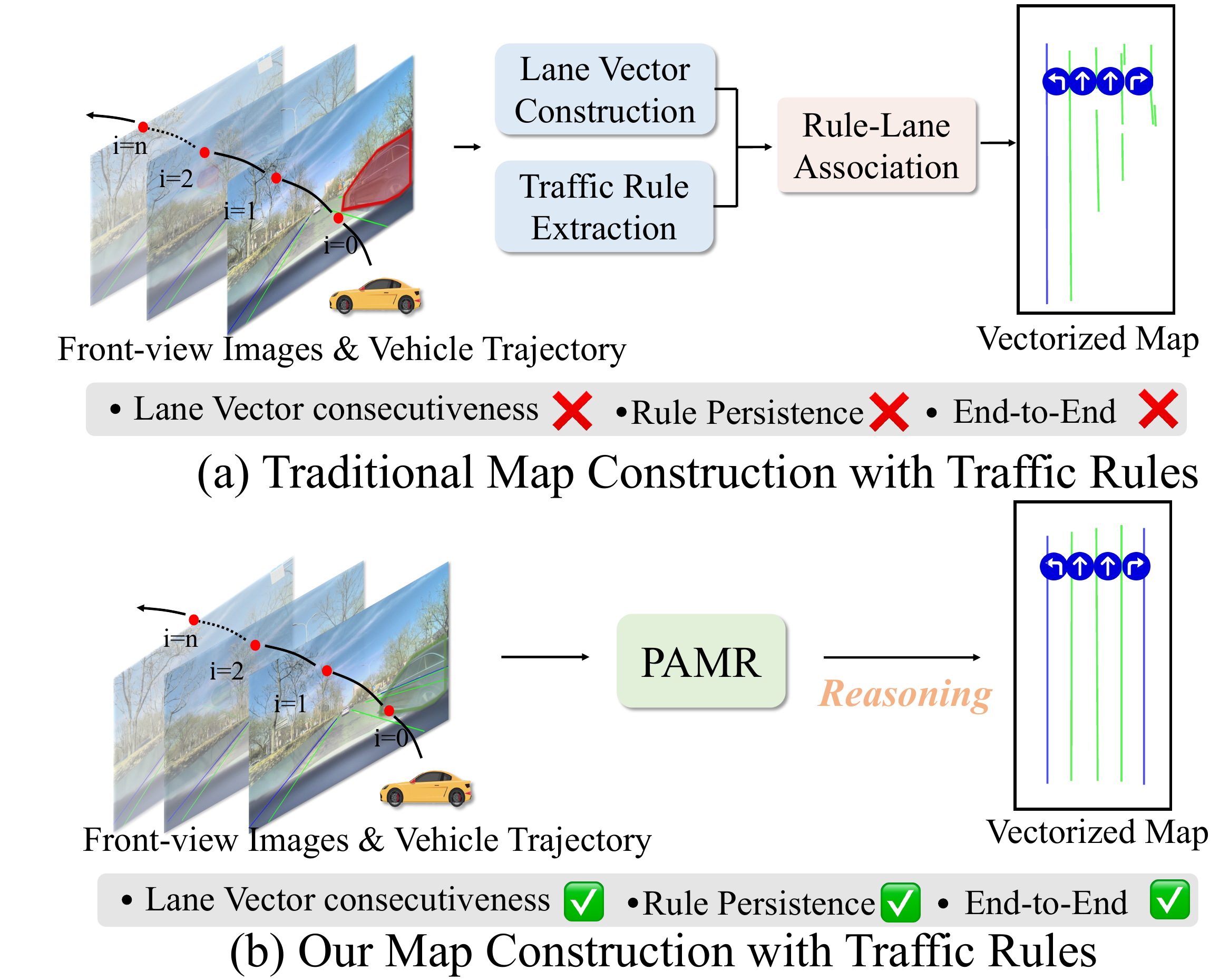} 
    \caption{Comparison between traditional pipeline and PAMR. (a) Traditional methods adopt a multi-stage pipeline: constructing lane vectors, extracting and associating traffic rules. This separated approach fails to maintain vector consecutiveness, rule persistence, and end-to-end learning. (b) Our PAMR framework performs autoregressive co-construction of lanes and rules, enabling sequential reasoning. Through joint construction, PAMR achieves continuous vectors, persistent rule awareness, and end-to-end integration of geometric and semantic information.}
    \label{fig:intro}
\end{figure}

Current methods fail to capture this persistent nature. The key challenge lies in traffic rules' persistent effectiveness: a sign's influence extends well beyond its visible range (Fig~\ref{fig:traffic}). This creates a complex dependency between road geometry and rules that existing systems cannot model. Traditional vector-only methods ~\cite{liao2023maptr, maptrv2} generate geometric vectors but remain   ``semantically blind." While some approaches~\cite{wang2023openlanev2} incorporate rules, they reduce rule assignment to simple classification, failing to maintain consistency or interpret signs' lasting impact. Even recent advances like MapDR~\cite{mapdr}, despite addressing rule understanding, rely on pre-constructed vectorized maps (Fig~\ref{fig:intro} (a)), preventing end-to-end geometry-semantic co-construction. This fundamental inability to model persistent effectiveness fragments the driving environment, failing to create the coherent representation necessary for rule-compliant navigation.

In this paper, we propose \textbf{PAMR} (\textbf{P}ersistent \textbf{A}utoregressive \textbf{M}apping with Traffic \textbf{R}ules), a novel framework that performs autoregressive co-construction of lane vectors and traffic rules from visual observations. Our approach integrates geometric reasoning with traffic semantics, enabling the model to infer lane-level rules and maintain their validity over time. This design mirrors human drivers' ability to apply traffic rules beyond immediate visibility, 
bridging the gap between instantaneous perception and rule-aware decision-making in autonomous systems. Rather than detecting isolated elements, PAMR ``narrates" the road scene by conditioning each map element on previously generated context. This sequential reasoning ensures consistency between lane structures and their governing rules, while enabling occluded lane inference through contextual understanding (Fig~\ref{fig:intro} (c)).

PAMR contains
two key mechanisms: \textbf{Map-Rule Co-Construction} for processing driving scenes in temporal segments, and \textbf{Map-Rule Cache} for seamless propagation of rules and geometry between segments. However, evaluating continuous and consistent map generation requires an appropriate benchmark. The original MapDR~\cite{mapdr} dataset, with its fragmented lane annotations, proves inadequate for this task. We therefore developed MapDRv2, featuring smooth and continuous lane geometries, to enable meaningful evaluation of our approach.

To sum up, our contributions are as follows:

\begin{itemize}

    \item
    We develop MapDRv2, a re-annotated dataset featuring continuous lane geometries, providing a more suitable benchmark for evaluating models that focus on generating consistent and continuous HD maps with traffic rules.
    
    \item 
    We propose PAMR, a novel framework that achieves persistent driving by the rules through autoregressive co-construction of lane vectors and traffic rules, addressing the fundamental challenge of maintaining rule effectiveness beyond immediate visibility.
    
    \item
    We introduce two key technical components: Map-Rule Co-Construction for processing driving scenes in temporal segments, and Map-Rule Cache for maintaining consistent rule propagation, enabling seamless integration of geometric and semantic information across extended driving sequences.

\end{itemize}

%% file: sec/2_related.tex
\section{Related Work}

\subsection{HD Map Construction}
Fueled by large-scale autonomous driving~\cite{Argoverse2, nuscenes2019} and traffic sign datasets~\cite{bstld, gtsrb, bdd100k, DTLD, tt100k}, HD map construction is rapidly evolving towards end-to-end solutions~\cite{li2024bevformer, zhang2024online, zhang2023online, yuan2025unimapgen}. Pioneering works like VectorMapNet~\cite{liu2022vectormapnet} and MapTR~\cite{liao2023maptr} established methods for sequential and permutation-equivalent vector prediction. However, despite architectural advances like PivotNet~\cite{ding2023pivotnet} for improving geometry, these methods remain ``semantically blind" to traffic rules.

Recent attempts to incorporate semantics are limited. OpenLane-V2~\cite{wang2023openlanev2} treats rule assignment as simple classification, while MapDR~\cite{mapdr} relies on pre-constructed vectors, breaking the end-to-end paradigm. A critical challenge therefore remains: modeling the persistent effectiveness of traffic rules beyond their visible range while ensuring geometric-semantic consistency. This gap motivates our work towards unified geometric-semantic mapping.

\subsection{MLLMs in Driving}
Driving's sequential nature requires continuous interpretation of semantics, yet even advanced sequential models like VectorMapNet~\cite{liu2022vectormapnet} and MapTR~\cite{liao2023maptr} largely neglect the temporal persistence of traffic rules. This is where Multimodal Large Language Models (MLLMs) show great promise, known for their success in sequential reasoning and contextual awareness~\cite{claude, Qwen-VL, DBLP:gpt4}.

Their potential for understanding complex driving scenarios is being actively explored~\cite{drivellm, Drivegpt4, sima2024drivelmdrivinggraphvisual, tian2024drivevlm, drivingwithllms, ding2024holistic, choudhary2024talk2bev, cao2025flame, zeng2025FSDrive, zeng2024driving, xie2025seqgrowgraph, wei2024editable, Xie_2025_ICCV, zeng2024driving, zeng2025FSDrive, zeng2025janusvln}. Particularly relevant is their ability to ``narrate" sequential information while maintaining long-term dependencies—a crucial capability for the persistent rule-aware mapping that current methods lack. This aligns with the trend towards unified architectures, offering a promising path to integrate deep geometric and semantic understanding in autonomous driving.

%% file: sec/4_E2E-ARM.tex
\section{Method}
\label{e2e-arm}

\begin{figure*}[htbp]
    \centering
    \includegraphics[width=0.85\linewidth]{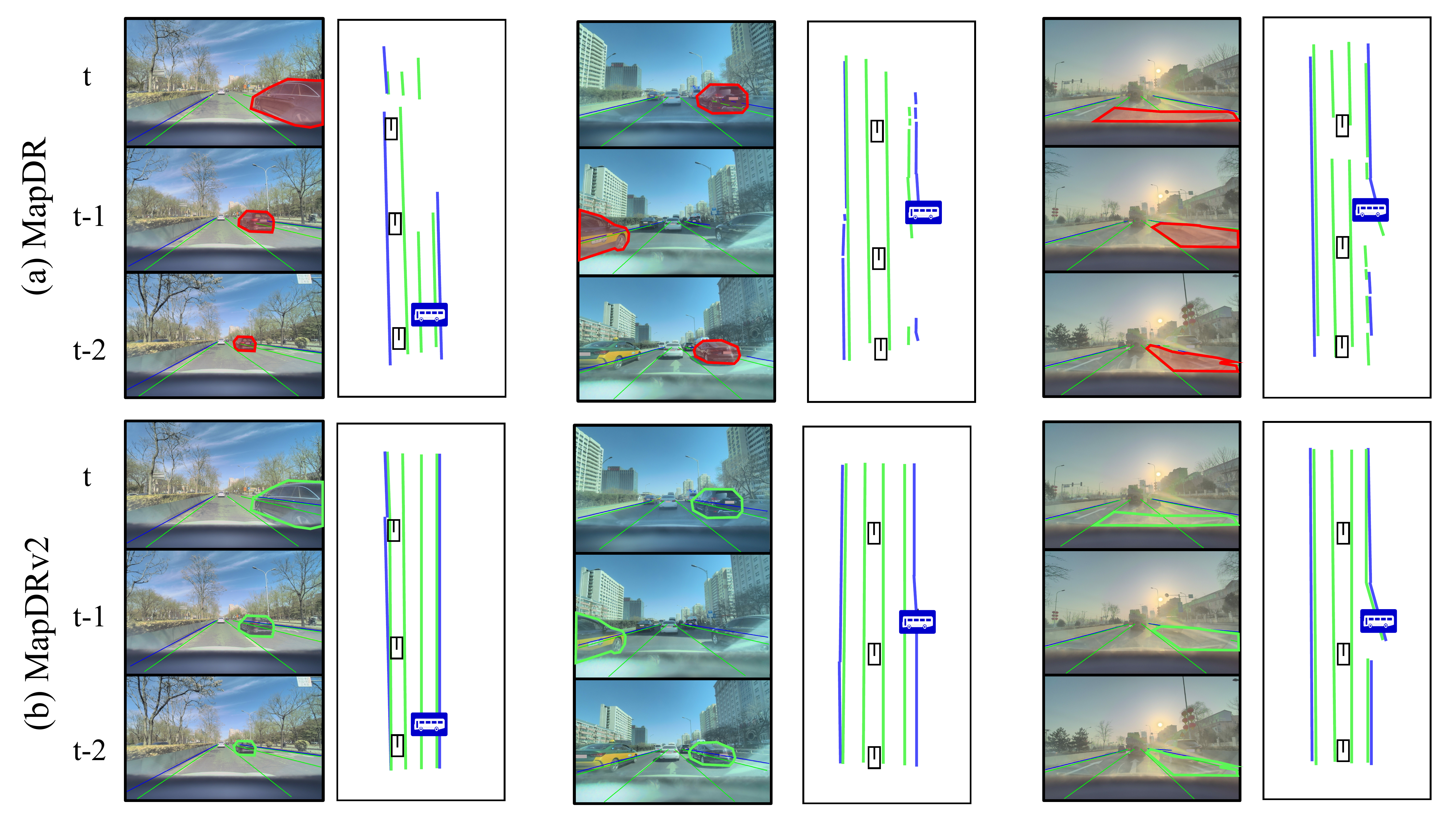} 
    \caption{Comparison with MapDR in occluded scenarios. For each example, the left three images show the input PV views, while the right images present the corresponding ground-truth HD maps. 
    Green lines and blue lines represent the dividers and borderlines, respectively.
    In the case of occlusion and glare, MapDR generates fragmented vector representations. In contrast, MapDRv2 provides more complete results.}
    
    \label{dataset}
\end{figure*}

\subsection{MapDRv2}
\label{sec:dataset}
Persistent driving by the rules necessitates a benchmark that upholds the same principles of continuity and coherence. A model's ability to generate continuous outputs is fundamentally determined by the integrity of its training data. While the original MapDR dataset was pioneering in its integration of traffic rules, its geometric annotations suffer from significant fragmentation, frequently exhibiting discontinuities in scenarios involving occlusion or complex topologies (see Fig~\ref{dataset}(a)). Consequently, this geometric fragmentation renders the dataset unsuitable for training or evaluating models whose primary objective is to generate a single, coherent representation of the road network.

To establish a robust foundation for this task, we performed a meticulous re-annotation of the lane vectors in the MapDR dataset. While preserving the original data scale and scenarios, we employed a human-in-the-loop methodology. This process involved projecting video frames into a unified BEV space, which enabled human annotators to leverage the full temporal context of each clip. By doing so, they could accurately extrapolate occluded lane segments based on contextual geometric cues and resolve topological ambiguities using logical priors. This procedure yielded a set of enhanced annotations characterized by smooth and continuous lane geometries that more faithfully represent real-world road structures, thereby providing a reliable ground truth for our restoration task (see Fig~\ref{dataset}(b)).

With this enhanced ground truth established, we developed a unified evaluation framework to holistically assess a model's performance. Our evaluation methodology is composed of two primary categories of metrics:

{\flushleft\textbf{Vectorized Lane Accuracy.}}
To assess predicted lane vectors $\hat{V}$ against ground-truth $V$, we implement topology-aware Intersection-over-Union (IoU ~\cite{zheng2020distance}) evaluation. For each lane polyline $\hat{l}_i \in \hat{V}$, we rasterize lanes into fixed-wide binary masks through polyline expansion. The IoU between each predicted mask $\hat{l}_i$ and ground-truth masks $\left \{ l_i \right \}_{i=1}^k$ is computed, with matches established when IoU $>$ 0.5. We define the vector accuracy metric $\mathcal{F}_{\text{vec}}$ as  Eq \ref{Eq:vec}:
\begin{equation}
\mathcal{F}_{\text{vec}} = \frac{1}{|\mathcal{M}|}\sum{(\hat{l}_i,l_j)\in\mathcal{M}}_\text{IoU}(\hat{l}_i, l_j)
\label{Eq:vec}
\end{equation}

{\flushleft\textbf{Holistic Mapping Accuracy.}}
We propose a single, unified metric, HMA, that evaluates the joint correctness of lane vectors, traffic rules, and their association. A predicted pair $(\hat{r}_i, \hat{l}_j)$ is considered true positive (TP) iff: (1) The predicted lane$\hat{l}_i$ accurately matches a ground-truth lane $l_j$. (2) The extracted rule $\hat{r}_i$ correctly matches a ground-truth rule $r_i$. (3) The association between the matched pair $(r, l)$ is valid in the ground truth.
The overall performance is then quantified by the holistic F1-score, $\mathcal{F}$, which harmonizes precision ($P$) and recall ($R$) calculated over these comprehensively validated true positives:
\begin{equation}
\mathcal{F} = \frac{2  \mathcal{P}\mathcal{R}}{\mathcal{P} + \mathcal{R}}
\label{Eq:overall}
\end{equation}

\begin{figure*}[h]
    \centering
    \includegraphics[width=0.95\linewidth]{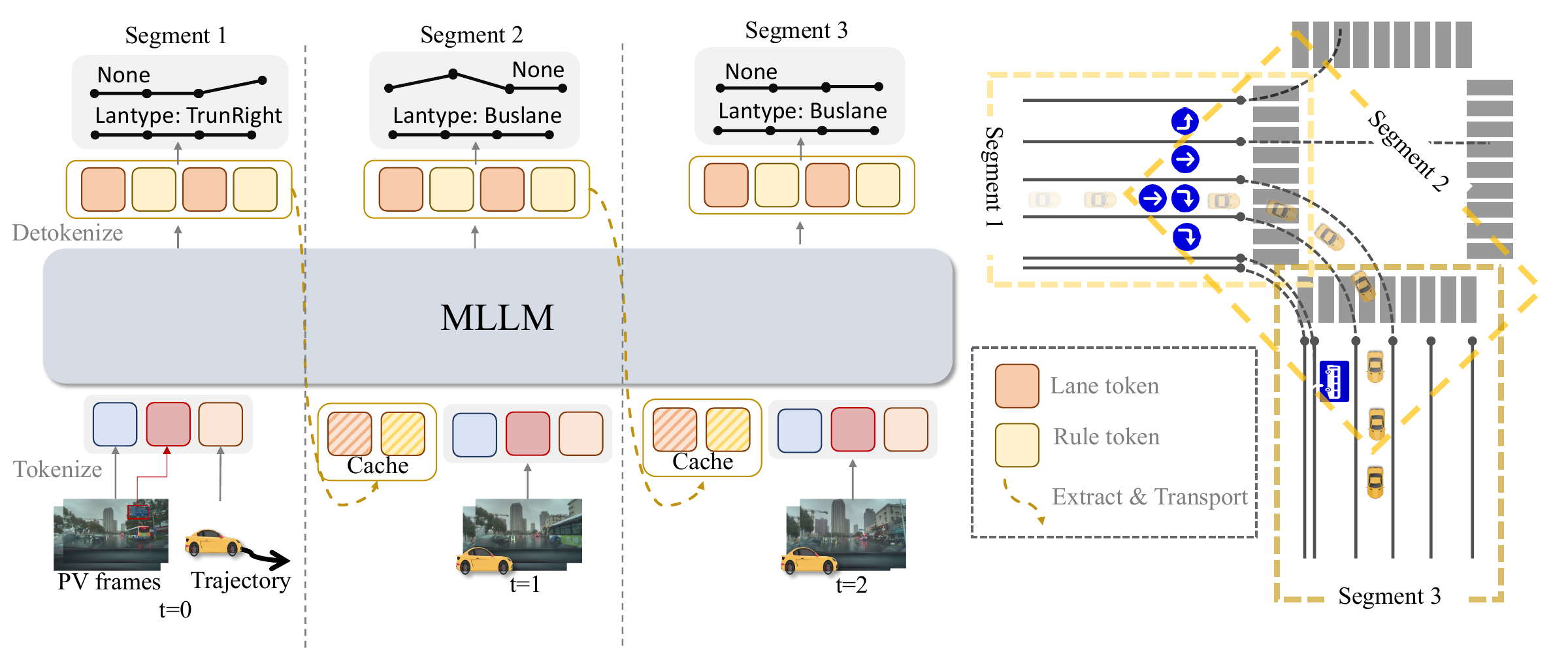} 
    \caption{Overview of the PAMR framework. \textbf{Left}: The sequential processing of map-rules, where each segment takes PV frames and trajectory as input, tokenizes them along with cache from previous segment (if any), and feeds them into MLLM. The MLLM outputs are then detokenized into lane vectors with associated rules. \textbf{Right}: Bird's-eye view visualization of the map-rule construction process, showing how information is propagated across consecutive segments through the caching mechanism. The cache ensures continuous rule awareness even as the vehicle moves forward, enabling consistent map-rule generation across the entire trajectory.}
    \label{fig:method_1}
\end{figure*}


\subsection{PAMR}

We propose \textbf{PAMR} (\textbf{P}ersistent \textbf{A}utoregressive \textbf{M}apping with Traffic \textbf{R}ules), an end-to-end framework for 
co-constructing lane vectors and traffic regulations. Our framework processes multimodal inputs and leverages MLLM~\cite{claude, DBLP:gpt4, geminiteam2024gemini, Qwen-VL} for comprehensive HD map construction. 

Specifically, our framework accepts a sequence of front-view images $I$ along with their corresponding vehicle trajectory $\mathcal{T}$ as input to generate a vectorized HD map, represented as a graph $G(V)$. Here $V\!=\!\{L,R\}$ consists of lane vectors $L\!=\!\{l_i\}_{i=1}^k$ and traffic rules $R\!=\!\{r_i\}_{i=1}^m$, This mapping process can be formally defined as Eq~\ref{eq:graph}:

\begin{equation}
    G(V) = f( I, \mathcal{T}, Prompt^*)
    \label{eq:graph}
\end{equation}

where $f$ denotes the MLLM, and $Prompt^{*}$ represents optional prompts that enable both continuous map-rule construction ($P_{con}$ in Sec~\ref{sec:patch}) and interactive ($P_{rule}$ in Sec~\ref{sec:prompt}). These components will be thoroughly discussed in the following sections.


\subsection{Map-Rule Co-Construction}

We propose an integrated approach that jointly generates lanes and rules within a map-rule framework, capturing both spatial and temporal contexts along the driving trajectory. At each timestamp $t$, we establish a segment $G_t$ centered at the ego vehicle's current position $(x_i,y_i)$, which encompasses a fixed spatial region of size  $W \times H$.  As illustrated in Fig~\ref{fig:method_1}, each segment integrates the current frame with a sequence of  $N$ historical frames to facilitate comprehensive vector and rule co-construction.

\textbf{Input Serialization.} 
Within each segment containing $N$ front-view images $\{I_t\}_{t=0}^{t=N}$, we first leverage a pre-trained visual backbone (e.g., Vision Transformer, ViT~\cite{dosovitskiy2020vit}) to extract high-level feature maps. These features are subsequently flattened and projected into a sequence of visual tokens, denoted as $T_{v}= \{IMG^j\}$. Meanwhile, we sample the vehicle's trajectory to obtain a series of ego-poses, where each pose is characterized by its 2D coordinates and heading angle $(x_t, y_t, \theta_t)$ in a local coordinate frame. To align with MLLM's discrete token processing paradigm, we perform quantization on this continuous data. Specifically, the coordinate and angle values are first normalized and discretized according to the current segment's dimensions, then mapped to unique tokens within the model's vocabulary, yielding a sequence of pose tokens $T_{pose}$. Finally, we interleave the token sequences from all modalities into a unified input stream $T_{in}=[T_{v0}, T_{pose0},..., T_{vN}, T_{poseN}]$.

\textbf{Output Serialization.}
Our framework implements a tightly coupled serialization scheme that integrates geometric information with semantic attributes at the generation level. Rather than producing geometries and rules separately, the MLLM is trained to generate a single, coherent sequence where each lane vector is immediately followed by its associated traffic rule.

The generation of a single lane vector $l_i$ follows a structured format: it begins with a start token \texttt{[lane]}, continues with a sequence of quantized 2D coordinate tokens representing its polyline, and terminates with an end token \texttt{[$\backslash$lane]}. 

Traffic regulations are encoded as key-value pairs linked to their corresponding geometric elements. Following a similar structure,  each rule begins with \texttt{[rule]}, contains pre-defined rule template tokens, and ends with \texttt{[$\backslash$rule]}. For lanes without associated rules, a special token \texttt{[None]} is used. The final outputs are organized as $output = \texttt{[lane]}  \dots \texttt{[$\backslash$lane]}\texttt{[rule]} \dots  \texttt{[$\backslash$rule]} \dots  $.



A deterministic parser processes this output string through a systematic procedure: first segmenting the sequence based on predefined delimiters, then de-quantizing the coordinate tokens into metric polylines, and finally associating each polyline with its corresponding rule attributes. This parsing process directly constructs the map graph $G(V)$.

\subsection{Map-Rule Cache}
\label{sec:patch}
As discussed in the Introduction, traffic rules exhibit persistent effects that extend well beyond their initial points of observation. Local segment consistency alone proves insufficient, as critical context may be lost when frames containing traffic signs move out of view. To address this limitation, we propose a map-rule cache that enables effective information propagation across consecutive segments.

As illustrated in Fig~\ref{fig:method_1}, our caching mechanism operates as follows: after processing each segment, we extract vector points near its boundary and project them into new map dimensions as cache data. These caches then serve as initialization states for subsequent segments, denoted as $p_{con}$. To ensure smooth transitions, we design caches with an overlapping region of length $\delta$. This process can be formalized:

\begin{equation}
     G_t(V) = f(X_t, T_t, P^{con}_{t-1})
    \label{eq:patch}
\end{equation}
Where $G_t$ represents the vectorized HD map within the current segment, $X_t$ denotes the PV images, and $\mathcal{T}_t$ indicates the trajectory. The term $\mathcal{P}^{con}_{t-1}$ represents the cache inherited from the previous segment. During concatenation, we resolve overlapping regions by retaining the prediction from the succeeding segment while discarding the corresponding region from the preceding one to ensure seamless integration.

This iterative process of propagation and concatenation enables the generation of arbitrarily long, continuous maps. Crucially, it ensures persistent rule awareness by maintaining historical context, even when the corresponding visual cues have moved beyond the current observation window.

\subsection{Interactive Prompt}
\label{sec:prompt}
Although driving scenarios typically involve multiple co-located traffic signs, and our PAMR model is fully capable of processing them simultaneously, we align our evaluation framework with the established MapDR protocol by focusing on individual signs. This selective evaluation is enabled by PAMR's interactive feature, which allows users or evaluation scripts to specify target traffic signs for map construction. This approach facilitates systematic, fine-grained assessment of the model's performance on each individual sign. Importantly, this single-sign focus during evaluation reflects a methodological choice rather than a model constraint - PAMR maintains the capability to process any combination of visible traffic signs concurrently.

To enable this interactivity, we introduce three distinct prompting strategies, formulated in Eq~\ref{eq:prompt_types}:
\begin{equation}
\mathcal{P}_{\text{rule}} \in \left\{
\begin{aligned}
&\texttt{[COORD]}\oplus (u_k^{(i)},v_k^{(i)}) \quad &&\text{(Coordinate)} \\
&\texttt{[BBOX]}\oplus b_k \quad &&\text{(Visual Highlight)} \\
&\texttt{[ROI]}\oplus X_k^{\text{ROI}} \quad &&\text{(Visual Rule)} \\
\end{aligned}
\right\},
\label{eq:prompt_types}
\end{equation}
where \texttt{[COORD]} leverages discretized 3D coordinates of traffic signs projected onto the PV as $(u_k^{(i)},v_k^{(i)})$, \texttt{[BBOX]} highlights the target traffic sign by drawing a bounding box $b_k$on the input image; and \texttt{[ROI]} extracts and encodes visual features from the traffic sign's region of interest.


%% file: sec/5_experiments.tex
\section{Experiments}

\begin{table*}[h]\small
    \centering
    \begin{tabular}{l|c|ccc|ccc}
    \toprule[2pt]
    \multirow{2}{*}{\centering Methods}  &
    \multicolumn{1}{c|}{Vec} & \multicolumn{3}{c|}{Rule Extract} & \multicolumn{3}{c}{\cellcolor{gray!25}{HMA}} \\
    \cmidrule(l){2-8}
    ~  & $\mathcal{F}_{vec}$ & $P_{\mathcal{H}}(\%)$ & $R_{\mathcal{H}}(\%)$ & $\mathcal{F}_{\mathcal{H}}$  & $P_{\mathcal{H}}(\%)$ & $R_{\mathcal{H}}(\%)$ & \cellcolor{gray!25}{$\mathcal{F}_{\mathcal{H}}$} \\
    \midrule[1pt]
    \textcolor{gray}{\textit{\small{single task}}} \\
    \textcolor{gray}{MapDR(VLE-MEE)} & /  & \textcolor{gray}{76.67} & \textcolor{gray}{74.54} & \textcolor{gray}{75.58} & \textcolor{gray}{63.35} & \textcolor{gray}{67.37} & \cellcolor{gray!25}{\textcolor{gray}{65.29}} \\
    \textcolor{gray}{MapDR(RuleVLM)}  & /  & \textcolor{gray}{89.28} & \textcolor{gray}{89.44} & \textcolor{gray}{89.3} & \textcolor{gray}{64.16} & \textcolor{gray}{64.25} & \cellcolor{gray!25}{\textcolor{gray}{64.20}}\\
    \midrule[1pt]
    \textit{\small{comprehensive task}} \\
    PAMR  & 0.46 & 84.52 & 82.22 & 83.39 & 71.32 & 69.33 & \cellcolor{gray!25}{\textbf{70.37}} \\
    \bottomrule[2pt]
    \end{tabular}
    \caption{Evaluation on MapDRv2 test set. Comparison of different methods across three metrics: \textbf{HMA} (holistic quality of rule extraction and contextual association), \textbf{Rule Extraction} (extracting key-value pairs from traffic sign rules), \textbf{Vec} (lane vector reconstruction accuracy). \textbf{Notably, MapDR directly provides target traffic signs, while PAMR must first identify the relevant signs from multiple candidates before performing rule extraction.}
    }
    \label{tab:main result}
\end{table*}

\subsection{Implementation Details}
{\flushleft\textbf{Dataset and Metric.}} We evaluate PAMR on MapDRv2, with metrics defined in Sec \ref{sec:dataset}. The dataset encompasses diverse scenarios, weather conditions, and traffic situations, comprising over $10$k traffic scene segments, $18$k driving rules, and $400$k images with $1960 \times 1240$ resolution. 
Each segment is set to  $224 \times 224$ ($W \times H$, corresponding to a $22.4m \times 22.4m$ real-world area), with an overlap ratio $\delta$ of $10\%$. The dataset is split into \textit{train} and \textit{test} sets with a $9:1$ ratio for evaluation.

{\flushleft\textbf{Training Strategy}} 
We use Qwen2-VL-2B~\cite{Qwen-VL} as our MLLMs model. For training, we set the batch size a 256, and the models are optimized using AdamW~\cite{adamw} with a weight decay of $0.1$. The learning rate is set to $2 \times 10^{-5}$ and a linear warm-up of $100$ steps. The training process comprises 20 epochs. $32$ NVIDIA H20 are used in total. In order to reduce the occupation of GPU memory, we uniformly sample each group of input PV images and retain less than 10 images and each image is resized to $644 \times 364$. Please refer to the supplementary materials for more training details.

\subsection{Main Result}

Table~\ref{tab:main result} presents a comprehensive evaluation of our proposed PAMR. While PAMR shows lower performance in rule extraction compared to RuleVLM~\cite{mapdr}, this is primarily because RuleVLM processes PV images containing \textbf{only} target rules, whereas PAMR needs to identify and evaluate specific signs from multiple traffic rules. Although this selective evaluation requirement affects PAMR's performance in single-rule metrics compared to MapDR, our method significantly outperforms both RuleVLM and VLE-MEE in joint rule extraction and association tasks. This superior performance in unified tasks demonstrates PAMR's effectiveness in joint vector-rule mapping, enabled by our co-construction approach that enables deep semantic understanding of rules rather than simple matching-based association.

\begin{table}[h]\small
    \centering
    \begin{tabular}{l|c|ccc}
    \toprule[2pt]
     \multirow{2}{*}{\centering Segment Size} &
    \multicolumn{1}{c|}{Vec}  & \multicolumn{3}{c}{HMA}   \\
    \cmidrule(l){2-5}
     ~  & $\mathcal{F}_{vec}$ & $P_{\mathcal{H}}(\%)$ & $R_{\mathcal{H}}(\%)$ & $\mathcal{F}_\mathcal{H}$\\
    \midrule[1pt]
     122 $\times$ 224  & \textbf{0.46} & 63.12 & 46.58 &  53.60 \\
     224 $\times$ 224  & \cellcolor{gray!25}\textbf{0.46} & \cellcolor{gray!25}\textbf{71.32} & \cellcolor{gray!25}\textbf{69.33}  &\cellcolor{gray!25}\textbf{70.37}     \\
     448 $\times$ 224 & 0.41  & 64.34 & 55.59 &  59.64 \\
    
    \bottomrule[2pt]
    \end{tabular}
    \caption{Segment Size. Performance comparison across different Segment widths with a fixed height. All configurations are designed to ensure complete coverage of local map segments.}
    \label{tab:patch}
      
\end{table}

\subsection{Ablation Study}

We conduct ablation studies to evaluate our key design choices. All experiments use identical settings except for the components under comparison, ensuring controlled evaluation. Configurations used in our final model are highlighted in \colorbox{gray!25}{gray}.

\begin{table}[t]\small
    \centering
        \centering
        
        \renewcommand{\arraystretch}{0.9} 
        \begin{tabular}{l|ccc}
            \toprule[2pt]
            \multirow{2}{*}{\centering Map-Rule Cache} & \multicolumn{3}{c}{HMA} \\
            \cmidrule(l){2-4}
             & $P_{R.E.}(\%)$ & $R_{R.E.}(\%)$ &  $\mathcal{F}_{rule}$  \\
            \midrule[1pt]
           w/o cache  & 41.24 & 38.73 & 39.93 \\
            w cache  & \cellcolor{gray!25}\textbf{71.32} & \cellcolor{gray!25}\textbf{69.33}  &\cellcolor{gray!25}\textbf{70.37}    \\
            \bottomrule[2pt]
        \end{tabular}
        \captionof{table}{Effectiveness Evaluation of Map-Rule Cache.}
        \label{tab:cache}
      
\end{table}

{\flushleft\textbf{Map-Rule Cache.}} 
We validate the importance of Map-Rule Cache through ablation studies, as shown in Table ~\ref{tab:cache}. Without the cache mechanism, all metrics show significant degradation. This significant degradation occurs because, in the absence of caching, rule association is limited to segments containing traffic signs, and rule propagation between segments is disrupted. Consequently, vector measurements become discontinuous and inconsistent across segments.

\begin{table}[t]\small
    \centering
        \centering
        
        \renewcommand{\arraystretch}{0.9} 
        \begin{tabular}{l|ccc}
            \toprule[2pt]
            \multirow{2}{*}{\centering Strategy} & \multicolumn{3}{c}{HMA} \\
            \cmidrule(l){2-4}
             & $P_{R.E.}(\%)$ & $R_{R.E.}(\%)$ &  $\mathcal{F}_{rule}$  \\
            \midrule[1pt]
            w/o prompt   & 44.52 & 39.30 & 41.74   \\
            \midrule[1pt]
            \texttt{[COORD]} & 58.52 & 53.99  & 56.16 \\
            \texttt{[BBOX]}  & 63.78 & 55.10 & 59.12  \\
            \texttt{[ROI]}   & \cellcolor{gray!25}\textbf{71.32} & \cellcolor{gray!25}\textbf{69.33}  &\cellcolor{gray!25}\textbf{70.37}    \\
            \bottomrule[2pt]
        \end{tabular}
        \captionof{table}{Interactive Prompt.}
        \label{tab:prompt}
      
\end{table}

%

{\flushleft\textbf{Segment Size.}} 
Our method employs local map-rule co-construction, where the segment dimensions critically influence the information density of local reconstruction. As shown in Table~\ref{tab:patch}, our size variation experiments reveal a clear trade-off: smaller segments excel in lane vector construction due to their focus on local details but compromise rule understanding due to limited context. Conversely, larger segments struggle with vector generation due to detail loss while introducing excessive noise that impairs rule interpretation. Medium-sized segments achieve optimal performance, maintaining sufficient detail for vector construction while capturing adequate context for rule understanding.

\begin{figure*}[t!]
\centering
\setlength{\tabcolsep}{1pt} 
\renewcommand{\arraystretch}{0.8} 
\includegraphics[width=0.95\linewidth]{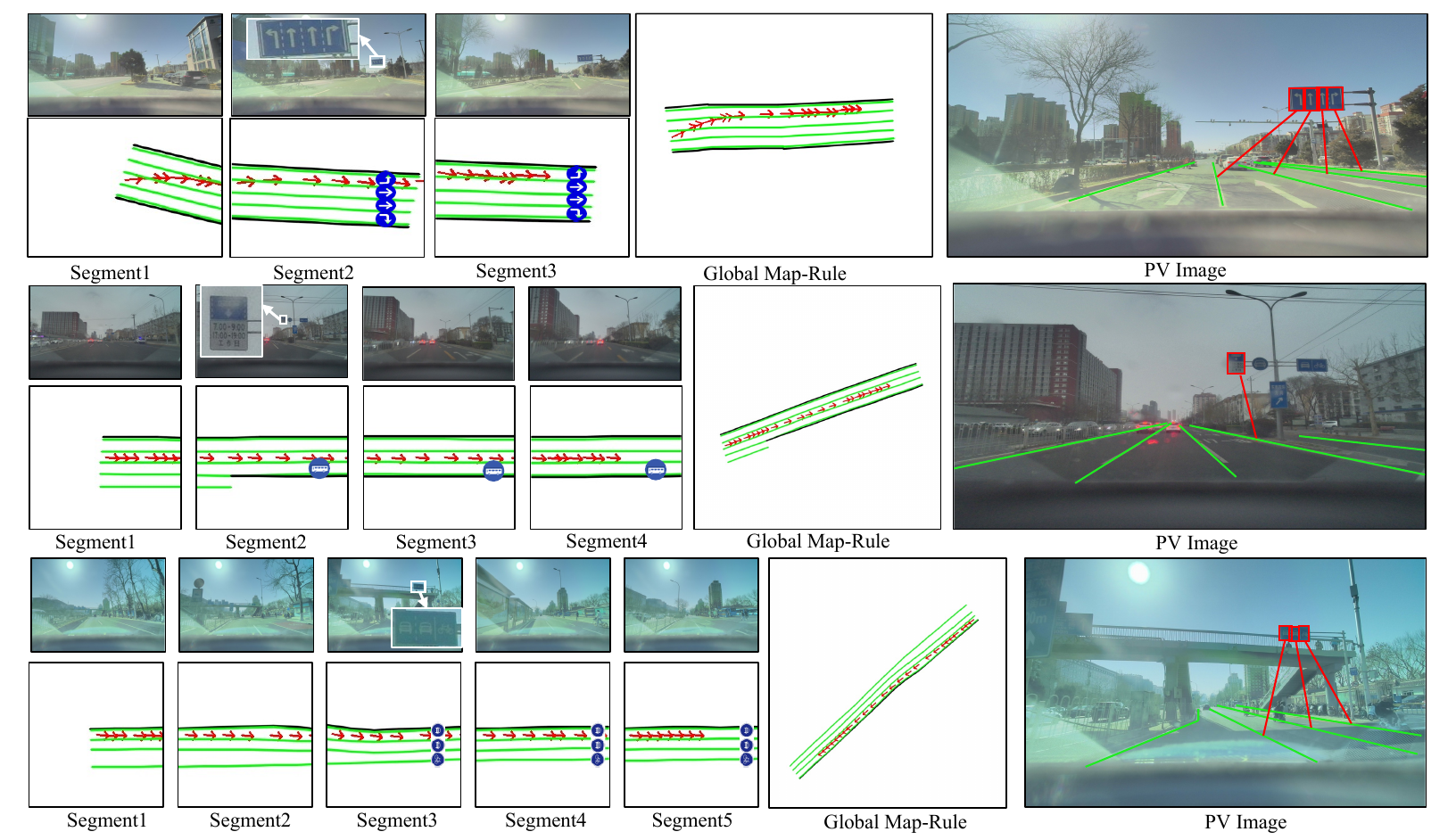}
\caption{Visualization of map-rule construction. Segment 1-5 demonstrate the sequential processing results within individual segments, while the HD map shows the final integrated output after segments are concatenation. The green lines represent the constructed lane vectors while the black lines indicate border lines, and the red arrow shows the vehicle trajectory within each segment.}
\label{fig:vis}
  
\end{figure*}

{\flushleft\textbf{Interactive Prompt.}} 
In Eq.~\ref{eq:prompt_types}, we propose three distinct prompting strategies $\mathcal{P}_{rule}$ to direct the model's attention to specific traffic signs. Table~\ref{tab:prompt} compares their effectiveness. The results demonstrate that directly incorporating visual ROI features achieves optimal performance. Alternative approaches, providing explicit coordinates or highlighting signs with bounding boxes, show reduced effectiveness. Without any prompting guidance, the model defaults to predicting all visible traffic signs, resulting in redundant outputs.

{\flushleft\textbf{Numbers of PV Images.}}
We investigate the impact of input sequence length by varying the number of PV images. As shown in Table \ref{tab:pv}, model performance improves with increasing input frames, suggesting that richer visual information enhances output quality. However, performance slightly degrades when exceeding 10 frames, indicating that excessive inputs may introduce redundant information that adversely affects model performance.

\begin{table}[t]\small
    \centering
        \centering
        \renewcommand{\arraystretch}{1.0} 
        \begin{tabular}{l|c|ccc}
        \toprule[2pt]
         \multirow{2}{*}{\centering Number}  & \multicolumn{1}{c|}{Vec} & \multicolumn{3}{c}{HMA}  \\
        \cmidrule(l){2-5}
        ~ & $\mathcal{F}_{vec}$ & $P_{\mathcal{H}}(\%)$ & $R_{\mathcal{H}}(\%)$ &  $\mathcal{F}_{\mathcal{H}}$ \\
        \midrule[1pt]
        5  & 0.42 & 67.15 & 62.35 &  64.66 \\
        10 & \cellcolor{gray!25}\textbf{0.46} & \cellcolor{gray!25}\textbf{71.32} & \cellcolor{gray!25}\textbf{69.33}  &\cellcolor{gray!25}\textbf{70.37}  \\
        15 & 0.43  & 68.43 & 61.33  & 64.68  \\
        \bottomrule[2pt]
        \end{tabular}
        \captionof{table}{Numbers of PV Images. Each experiment uniformly samples frames from the complete image sequence.}
        \label{tab:pv}
      
\end{table}

{\flushleft\textbf{Overlapping Region.}}
Table ~\ref{tab:overlap} presents ablation studies on the overlap region size, with map-rule dimensions held constant across experiments. Here, $\delta \%$ represents the ratio of overlap width to map-rule width. The results show that both insufficient and excessive overlap adversely affect vector construction and rule understanding performance.

\begin{table}[]\small
        \centering        
        \renewcommand{\arraystretch}{1.0} 
        \begin{tabular}{l|c|ccc}
        \toprule[2pt]
         \multirow{2}{*}{\centering $\delta \%$}  & \multicolumn{1}{c|}{Vec} & \multicolumn{3}{c}{HMA}  \\
        \cmidrule(l){2-5}
        ~ & $\mathcal{F}_{vec}$ & $P_{\mathcal{H}}(\%)$ & $R_{\mathcal{H}}(\%)$ &  $\mathcal{F}_{\mathcal{H}}$ \\
        \midrule[1pt]
        5 &  0.37  & 59.50 & 56.96 & 58.20 \\
        10 & \cellcolor{gray!25}\textbf{0.46} & \cellcolor{gray!25}\textbf{71.32} & \cellcolor{gray!25}\textbf{69.33}  &\cellcolor{gray!25}\textbf{70.37}   \\
        15 & 0.39 & 59.96 & 45.63 & 51.82  \\
        \bottomrule[2pt]
        \end{tabular}
        \captionof{table}{Overlapping Region. $\delta \%$ represents the proportion of segments that overlap.}
        \label{tab:overlap}
      
\end{table}


\subsection{Visualization}
Our framework demonstrates robust performance across various challenging scenarios, as illustrated in Fig~\ref{fig:vis}. The map-rule mechanism effectively addresses several key challenges in HD map construction. First, it maintains consistency during lane-changing scenarios and generates smooth, accurate curved lane vectors, demonstrating precise vector modeling capabilities. Second, our cache mechanism successfully preserves rule awareness even when traffic signs move beyond the current map-rule. Additionally, the model exhibits strong semantic understanding by accurately associating multiple rules from individual traffic signs with corresponding lanes.

%

%% file: sec/6_conclusion.tex
\section{Conclusion}
We propose PAMR, a novel framework for persistent rule-aware HD map construction. By integrating map-rule co-construction with a cache mechanism, PAMR successfully maintains rule awareness across extended driving sequences, addressing a critical challenge in autonomous navigation. Additionally, we introduce MapDRv2, featuring re-annotated continuous vector labels, to enable comprehensive evaluation of rule-aware mapping systems. Extensive experiments on MapDRv2 demonstrate PAMR's effectiveness in joint vector-rule mapping and consistent rule propagation.

\section{Limitations}
Our current evaluation is limited to the MapDRv2 benchmark. Although we envision PAMR as a universal plug-and-play solution for rule-aware HD mapping, its broader applicability requires validation. Future research will focus on extending PAMR to diverse mapping scenarios and benchmarks to verify its effectiveness as a general-purpose solution.